\documentclass[conference, letterpaper]{IEEEtran}
\IEEEoverridecommandlockouts
\usepackage{cite}
\usepackage[ruled]{algorithm2e}
\usepackage{amsmath,amssymb,amsfonts}
\usepackage{algorithmic}
\usepackage{array}
\usepackage{balance}
\usepackage{cellspace}
\usepackage{graphicx}
\usepackage{hhline}
\usepackage{subfigure}
\usepackage{makecell}
\usepackage{multirow}
\usepackage{textcomp}
\usepackage[table]{xcolor}
\def\BibTeX{{\rm B\kern-.05em{\sc i\kern-.025em b}\kern-.08em
\usepackage[left=0.75in, right=0.75in, top=0.75in, bottom=0.75in]{geometry}
    T\kern-.1667em\lower.7ex\hbox{E}\kern-.125emX}}
\usepackage[left=0.625in, right=0.625in, top=0.75in, bottom=1.05in]{geometry}
\title{Optimal Signal Decomposition-based Multi-Stage Learning for Battery Health Estimation\\
\thanks{This work was supported in part by A*STAR under its MTC Programmatic (Award M23L9b0052), SIT’s Ignition Grant (STEM) (Grant ID: IG (S) 2/2023 – 792), MTC Individual Research Grants (IRG) (Award M23M6c0113), and the National Research Foundation, Singapore and Infocomm Media Development Authority under its Future Communications
Research \& Development Programme (Grant FCP-SIT-TG-2022-007).}
}

\makeatletter
\newcommand{\linebreakand}{%
  \end{@IEEEauthorhalign}
  \hfill\mbox{}\par
\mbox{}\hfill\begin{@IEEEauthorhalign}
}
\makeatother

\author{
\IEEEauthorblockN{Vijay Babu Pamshetti}
\IEEEauthorblockA{
\textit{Singapore Institute of Technology, Chaitanya Bharathi Institute of Technology}\\
\textit{Singapore, 828608}\\
vijaybabu.pamshetti@singaporetech.edu.sg, vijaybabup$\_$eee@cbit.ac.in}
\and
\IEEEauthorblockN{Wei Zhang$^*$\thanks{$^*$ Corresponding author. Email: wei.zhang@singaporetech.edu.sg.}}
\IEEEauthorblockA{
\textit{Singapore Institute of Technology}\\
\textit{Singapore, 828608}\\
wei.zhang@singaporetech.edu.sg}
\linebreakand
\IEEEauthorblockN{King Jet Tseng}
\IEEEauthorblockA{
\textit{Singapore Institute of Technology}\\
\textit{Singapore, 828608}\\
kingjet.tseng@singaporetech.edu.sg}
\and
\IEEEauthorblockN{Bor Kiat Ng}
\IEEEauthorblockA{
\textit{Singapore Institute of Technology}\\
\textit{Singapore, 828608}\\
borkiat.ng@singaporetech.edu.sg}
\and
\IEEEauthorblockN{Qingyu Yan}
\IEEEauthorblockA{
\textit{Nanyang Technological University}
\\\textit{Singapore, 639798}
\\alexyan@ntu.edu.sg
}
}

\begin{document}
\bstctlcite{IEEEexample:BSTcontrol}

\maketitle

\begin{abstract}
Battery health estimation is fundamental to ensure battery safety and reduce cost. However, achieving accurate estimation has been challenging due to the batteries' complex nonlinear aging patterns and capacity regeneration phenomena. In this paper, we propose OSL, an \underline{o}ptimal \underline{s}ignal decomposition-based multi-stage machine \underline{l}earning for battery health estimation. OSL treats battery signals optimally. It uses optimized variational mode decomposition to extract decomposed signals capturing different frequency bands of the original battery signals. It also incorporates a multi-stage learning process to analyze both spatial and temporal battery features effectively.  An experimental study is conducted with a public battery aging dataset. OSL demonstrates exceptional performance with a mean error of just 0.26\%. It significantly outperforms comparison algorithms, both those without and those with suboptimal signal decomposition and analysis. OSL considers practical battery challenges and can be integrated into real-world battery management systems, offering a good impact on battery monitoring and optimization.
\end{abstract}


\section{Introduction}
Batteries have revolutionized modern technology with applications spanning portable electronics, electric vehicles, energy storage systems, etc. Li-ion battery is one of the most popular battery types thanks to its rapid charging capabilities, low self-discharge rates, and longevity. The demand for Li-ion batteries is expected to reach 4.7 TWh by 2030 with an annual growth of over 30\% \cite{fleischmann2023battery}. Despite the surging demand, batteries' complex electrochemical systems present significant operational challenges, and battery health shall be well monitored and predicted to ensure battery safety, reduce costs, and extend lifespan \cite{10253731,Zhao2024Practical}. A common battery health indicator is the state of health (SoH). It measures the ratio of the battery's remaining capacity to its rated capacity, e.g., when the battery is brand-new. Research efforts have been made to estimate and predict SoH accurately, yet challenges remain.

Most recent research works in this domain are machine learning (ML) based. For example, Zhang et al. \cite{10253731} introduce a self-attention graph pooling convolutional network (SAGPCN), which optimizes network structure with node characteristics and graph topology for SoH estimation. Jiang et al. \cite{Jiang08082024} utilize bidirectional long short-term memory (LSTM) networks enhanced by attention mechanisms for complex temporal feature extraction. Ansari et al. \cite{ANSARI2023109198} and Ma et al. \cite{MA2022104750} propose SoH neural networks with hyper-parameters optimized by meta-heuristics. A common feature of these existing solutions is that battery measurements are processed as time-series data which is often the default data format for battery instrumentation sensors. However, time-series analysis alone faces significant challenges in handling complex battery data characteristics, particularly capacity regeneration and measurement noises.

Recently, researchers have proposed to complement time-dimension battery data with frequency-relevant signals. Given the time-based battery signals, existing signal decomposition methods can decompose such signals into several signals for different frequency bands. One of the early methods is empirical mode decomposition (EMD). Liu et al. \cite{9040661} apply EMD to decompose battery signals and process the decomposed signals with LSTM for battery capacity estimation. EMD and its variants face limitations with complex nonlinear sequences, often resulting in suboptimal decomposition. Thus, variational mode decomposition (VMD) \cite{6655981} has emerged as a new and superior alternative to EMD in battery health estimation with its enhanced resilience to noise and robust mathematical foundation. Ding et al. \cite{9758685} and Chen et al. \cite{CHEN2024113388} integrate VMD with gated recurrent units and Transformer, respectively. VMD's effectiveness in signal decomposition relies on a few parameters, which are often optimized with meta-heuristics, e.g., cuckoo search algorithm (CSA) \cite{9758685}. Despite these advances, challenges persist in optimizing the integration of signal decomposition and ML. It seems challenging to capture the battery dynamics complexity sufficiently with a single ML. 

In this paper, we propose OSL, an \underline{o}ptimal \underline{s}ignal decomposition-based multi-stage \underline{l}earning for estimating battery SoH. The optimity of OSL lies in its usage of VMD with parameters optimized by particle swarm optimization (PSO) \cite{Eberhart1995}, as well as two-stage MLs with convolutional neural network (CNN) and LSTM. Given the collected battery signals, our PSO customizes the optimal parameters of VMD for the battery. VMD with its battery-specific optimal parameters decomposes the battery signals into several intrinsic mode functions (IMFs), each corresponding to a frequency band. We expect the high-frequency band to capture battery noises and rapid variations and the low-frequency band to represent trends such as long-term capacity decline. IMFs together offer a comprehensive understanding of the battery behavior and improve the effectiveness of the following learning process. CNN as the first stage of learning aims to extract spatial features from the IMFs and LSTM in the second stage handles temporal information. Finally, OSL leverages VMD's effective signal decomposition and CNN-LSTM's in-depth signal processing to estimate battery SoH accurately. Specifically, we have the following main contributions in this paper.
\begin{itemize}
    \item We proposed OSL which integrates signal decomposition and multi-stage ML for optimal battery health estimation.
    \item We optimized OSL and observed that PSO converges to the optimal fast for VMD parameter optimization.
    \item We conducted an experimental study and demonstrated that OSL achieves just 0.26\% mean errors and performs the best compared to existing algorithms.
\end{itemize}

The rest of the paper is organized as follows. We present OSL for battery health estimation in Section \ref{sec:osl}. In Section \ref{sec:exp}, we conduct an experimental study and present results and discussions. Finally, we conclude this paper in Section \ref{sec:conclusion}.

\section{OSL Methodology}
\label{sec:osl}
We present the methodology and technical details of our proposed OSL for battery health estimation in this section.

\subsection{OSL Input and Output}
\begin{figure}[t]
    \centering
    \includegraphics[width=0.48\textwidth]{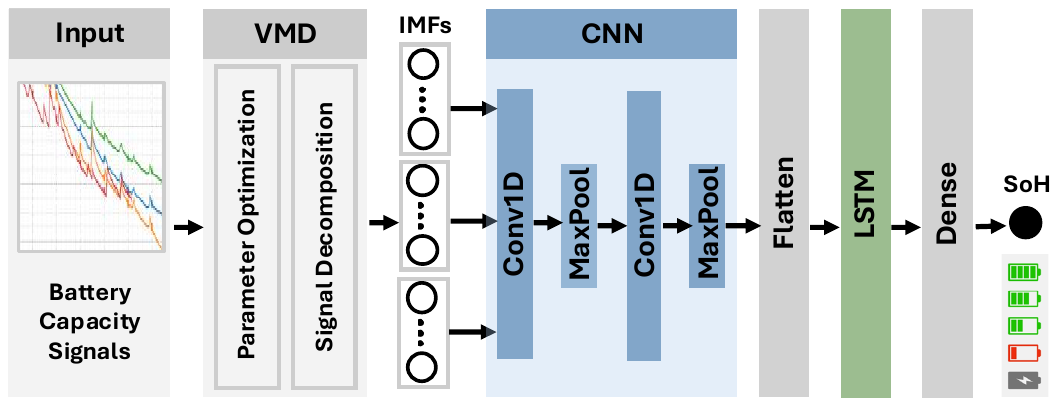} 
    \caption{An illustration of the proposed OSL. The time-based battery signals are decomposed into the signals for different frequency bands by VMD. The decomposed signals are processed by CNN followed by LSTM for effective battery data analysis and accurate battery health estimation.}
    \label{fig:osl}
\end{figure}

An illustration of the system architecture of OSL is shown in Fig. \ref{fig:osl}. OSL aims to correlate battery signals with battery health. Let $f(\cdot)$ be the input battery signals. Battery dynamics in practice can be measured for several aspects, e.g., capacity, current, voltage, and temperature. In this paper, we aim to understand the strengths and limitations of signal processing and ML for batteries, and we prefer a simple model with capacity as the only input to facilitate the understanding. Thus, $f(\cdot)$ in this paper is the battery capacity degradation curve in the time dimension. OSL processes and analyzes the signal $f(\cdot)$ up to time slot $t$ and estimates the battery health at time $t$. OSL can be easily configured to forecast battery health in the future and this can be discussed in the extended version of this paper. Specifically, let $f(t-T^\text{in}),\ldots,f(t-1)$ be the input of OSL at time $t$ where $T^\text{in}$ is the length of the input window. The output of OSL is the estimated SoH $\hat{y}^\text{SoH}(t)$ for time $t$ and let $y^\text{SoH}(t)$ be the ground-truth. OSL aims to approximate such a ground-truth with $\hat{y}^\text{SoH}(t)$ as close as possible.

\subsection{Signal Decomposition with VMD}
Unlike most ML-based battery health models, OSL complements ML with signal processing to enrich the input features for learning enhancement. Most measured battery signals are time-based, e.g., voltage time-series, and ML algorithms naturally process battery signals and model battery health estimation as a time-series problem. We argue that time is only one of the dimensions of battery signals and we incorporate frequency-based knowledge in OSL by applying signal decomposition. Among various signal decomposition methods, we choose VMD which effectively decomposes time-series data into a set of IMFs. Each IMF is characterized by a frequency band, referred to as bandwidth. Essentially, VMD has two objectives: 1) bandwidth shall be compact with IMFs sharing minimal overlap; 2) the summation of all IMFs shall reconstruct the original signal $f(\cdot)$. Thus, VMD can be formulated as the following  constrained variational problem,
\begin{equation}
\begin{aligned}
\min_{u_k,\omega_k} & \left\{ \sum\nolimits_{k=1}^{K} \left\| \frac{\partial}{\partial t} \left[ \Big(\delta(t) + \frac{j}{\pi t}\Big) * u_k(t) \right] e^{-j \omega_k t} \right\|_2^2 \right\}, \\
& \quad\quad \text{subject to } \sum\nolimits_{k=1}^{K} u_k(t) = f(t),
\end{aligned}
\label{eq:vmd-prob}
\end{equation}
where $K$ is the number of resulting IMFs and $u_k$ is the $k$-th IMF for $1\leq k\leq K$. Each IMF $u_k$ covers a frequency band and let $\omega_k$ be the central frequency of the band. The $\sum$ operation calculates the total bandwidth of all IMFs and VMD aims to minimize it so the IMFs are compact. To calculate each $u_k$'s bandwidth, the Hilbert kernel $\delta(t) + \frac{j}{\pi t}$ is used where $\delta(t)$ represents the original signal and $\frac{j}{\pi t}$ is the Hilbert transform of the signal. Signal $u_k$ is convoluted, operation $*$, with the Hilbert kernel to produce its analytic representation. Term $e^{-j \omega_k t}$ is used for frequency shift such that $u_k$ is shifted to the near-zero frequency band by subtracting center frequency $\omega_k$. Then, we employ squared L2-norm $\left\|\cdot \right\|_2^2$ to measure the signal magnitude after transformation. For simplicity, let $b_k= \frac{\partial}{\partial t} \left[ \Big(\delta(t) + \frac{j}{\pi t}\Big) * u_k(t) \right] e^{-j \omega_k t}$ be $u_k$'s bandwidth in the following parts. A constraint for minimizing the total bandwidth is $\sum\nolimits_{k=1}^{K} u_k(t) = f(t)$, meaning that the summation of the IMFs reconstructs $f(\cdot)$.

To solve the constrained optimization problem outlined in Eq. (\ref{eq:vmd-prob}), an augmented Lagrangian is introduced as,
\begin{equation}
\begin{aligned}
\mathcal{L}(u_k, \omega_k, \lambda) =& \alpha \sum_{k=1}^{K} \left\| b_k \right\|_2^2 + \left\| f(t) - \sum_{k=1}^{K} u_k(t) \right\|_2^2 \\
&+ \langle \lambda(t), f(t) - \sum_{k=1}^{K} u_k(t) \rangle,
\end{aligned}
\label{eq:vmd-lagrangian}
\end{equation}
which has three terms. The first term captures the compactness of the IMFs as described above. The second term reflects the constraint in Eq. (\ref{eq:vmd-prob}) for signal reconstruction fidelity. In reality, the two objectives of VMD can be contradictory, as minimizing frequency bandwidth may exclude certain signals and reduce the reconstruction fidelity. So, the two objectives are balanced with a parameter $\alpha$. The second term imposes a soft constraint that may not be sufficient for achieving reconstruction fidelity, e.g., when the first term is too dominant. Thus the third term with the Lagrange multiplier is introduced to enforce the constraint and the impact of this term can be controlled by $\lambda$. Finally, $\mathcal{L}(\cdot)$ in Eq. (\ref{eq:vmd-lagrangian}) can be optimized iteratively to search the optimal $u_k$ and $\omega_k$ with updated $\lambda$ reflecting the reconstruction error of each iteration.    

\subsection{VMD Parameters Optimization}
In the above introduced VMD optimization, there are several parameters that affect the optimization results. Two important parameters are $K$, for the number of IMFs, and $\alpha$, for balancing IMF compactness and signal reconstruction fidelity. The optimal parameters are often not too small neither too large. For example, a small $K$ may result in inadequate decomposition with poorly modeled patterns for batteries. Conversely, a large $K$ may lead to over-decomposition and ML algorithms can be biased by the decomposed components with limited relevance with battery health. A challenging fact is that the optimal parameters are not fixed and they vary for different applications. Thus it is important to optimize the parameters for our battery analytics application.

Several meta-heuristics have been employed to find the optimal $K$ and $\alpha$ \cite{CHEN2024113388}. In this paper, we choose PSO, an established and widely used algorithm that is widely used in many applications. PSO is effective in optimization because of its distinct features such as memory retention, collective learning, and efficient balance between global exploration and local exploitation. In this study, we configure PSO with a two-dimensional particle swarm formulation to simultaneously determine optimal $K$ and $\alpha$. The algorithm iteratively updates each particle's position and velocity. The effectiveness of PSO depends on a number of parameters, e.g., inertia weight and learning rates. An appropriate balance of these parameters accelerates convergence and improves the likelihood of finding the global optimal. To enhance performance, a linearly varying inertia weight is applied to balance global and local search capabilities. The optimal learning rates are taken from \cite{Eberhart1995}. 

During PSO, the effectiveness of each particle swarm needs to be quantified and we introduce a fitness function as the envelope entropy given by,
\begin{equation}
H = \sum_{k=1}^K\Big(-\sum_{t} P_k(t) \log P_k(t)\Big),~P_k(t) = \frac{e_k(t)}{\sum_{t} e_k(t)},
\end{equation}
where $e_k(t)$ is the mean envelop of IMF $u_k$, calculated by $\big(e_k^\text{up}(t) + e_k^\text{low}(t)\big)/2$ where the two terms are the upper and lower envelops, respectively. The upper and lower envelops are derived by identifying a sequence of local maxima and minima and approximating the local extrema with smooth curves, i.e., $e_k^\text{up}(t)$ and $e_k^\text{low}(t)$. Accordingly, $P_k(t)$ serves as the probability distribution of the envelop amplitudes and $H$ captures the total Shannon entropy of $K$ IMFs.

We are the first team to use PSO for VMD parameter optimization for battery health estimation. However, our team does not intend to claim such a novelty. This is based on our observations that the impact of using different meta-heuristics on optimizing $K$ and $\alpha$ is very minimal and several algorithms produce the same optimal parameters. PSO was chosen here for its optimal performance, fast convergence, and abundant library support. We suggest future research efforts to prioritize signal processing and ML algorithms for battery analytics.

\subsection{Multi-Stage ML for OSL}
VMD based on its optimal $K$ and $\alpha$ produces $K$ IMFs, which are used to train ML models and estimate battery SoH. OSL employs the multi-stage CNN-LSTM to analyze the IMFs. To estimate the SoH at time $t$, CNN is employed in the first stage and CNN's input is the three IMFs from time $t-T^\text{in}$ to time $t-1$, i.e., three sequences each of length $T^\text{in}$. The input will be processed by a convolutional layer \texttt{Conv1D} first to extract spatial patterns related to battery health. Then a max pooling layer \texttt{MaxPool} is configured to reduce the dimensionality of the extracted features from the first \texttt{Conv1D}. The second \texttt{Conv1D} is configured to further extract more complex spatial features that are expected to be more correlated with battery health. Similarly, another \texttt{MaxPool} is introduced for dimensionality reduction. Both convolutional layers use \texttt{ReLU} as the activation function. The last step of the CNN stage is to map the multi-dimensional features into a one-dimensional vector for each time slot with a flatten layer \texttt{Flatten}. Then the vectors are processed by LSTM with a number of LSTM cells and \texttt{ReLU} activation in the second stage, to extract temporal dependencies from the vectors. Finally, our model concludes with a fully connected layer \texttt{Dense} that synthesizes the LSTM's one-dimensional output vector to generate the final prediction for the battery SoH. Overall, we expect comprehensive battery feature extraction and analysis with the two-stage CNN-LSTM to enhance learning capabilities with both spatial and temporal features.

\section{Experimental Study}
\label{sec:exp}
We conduct an experimental study and present the results and discussions in this section.

\subsection{Dataset}
\begin{figure}[t]
    \centering
    \includegraphics[width=0.48\textwidth]{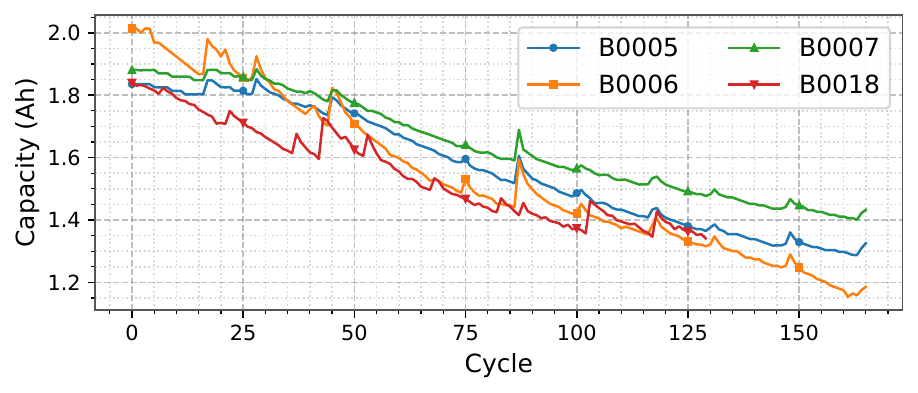} \vspace*{-2em}
    \caption{Capacity degradation curves of the four batteries in the NASA dataset. Capacity decreases in general but is not monotonic.}
    \label{fig:nasa}
\end{figure}

To demonstrate the effectiveness of the proposed approach, we use the dataset collected by NASA Ames Advanced Development Center \cite{saha2007battery}. The dataset includes four 18650 LiCoO$_2$ batteries, labeled as B0005, B0006, B0007, and B0018, each with a rated capacity of 2 Ah. The batteries were run through three operational profiles, charge, discharge, and impedance at room temperature. Charging was carried out in a constant current mode at 1.5 A until a charging cutoff voltage of 4.2 V, and then continued in a constant voltage model until the charge current dropped to 20 mA. Discharge follows a constant current at 2 A until the voltage decreases to 2.7, 2.5, 2.2, and 2.5 V, respectively for the four batteries. The capacity degradation of the batteries is visualized in Fig. \ref{fig:nasa}. We can see that the repeated charge and discharge cycles result in battery aging and capacity degradation, though the degradation is not monotonic. The data collection was stopped when the batteries reached end-of-life, which corresponds to the cycle when a battery's capacity is 30\% lower than its rated capacity.

\begin{figure}
    \centering
    \includegraphics[width=0.48\textwidth]{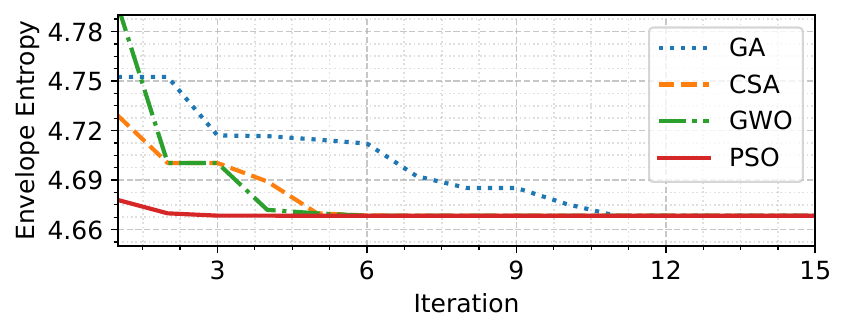} \vspace*{-2em}
    \caption{The convergence of the fitness value based on envelop entropy with different algorithms for battery B0005. All algorithms converge to the same optimal fitness value and PSO converges the fastest.}
    \label{fig:pso}
\end{figure}

There are several features available in the dataset such as the measured voltage, current, and temperature. In this paper, we follow a simplified model, where the input consists of the battery capacity degradation data only. The output of the model is a battery's SoH, and note that SoH is not part of the input.

\subsection{Evaluation Criteria}
Battery health is expected to be estimated accurately. In this paper, the estimation performance is measured with popular metrics, including mean absolute error (MAE), mean absolute percentage error (MAPE), and root mean square error (RMSE). A small value means high estimation accuracy.
% and the calculation equations are,
% \begin{equation}
%     \text{MAE} = \frac{1}{n} \sum\nolimits_{t=1}^{n} |\hat{y}^\text{SoH}(t) - y^\text{SoH}(t)|,
% \end{equation}
% \begin{equation}
% \text{MAPE} = 100\% \cdot \frac{1}{n} \sum\nolimits_{t=1}^{n} \left| \frac{\hat{y}^\text{SoH}(t) - y^\text{SoH}(t)}{y^\text{SoH}(t)} \right|,
% \end{equation}
% \begin{equation}
%     \text{RMSE} = \sqrt{\frac{1}{n}\sum\nolimits_{j=1}^{n} (\hat{y}^\text{SoH}(t) - y^\text{SoH}(t))^2},
% \end{equation}
% where $n$ represents the number of cycles for capacity estimation, $\hat{y}^\text{SoH}(t)$ is the predicted SoH for cycle $t$, and $y^\text{SoH}(t)$ is the SoH ground-truth of the cycle. A good SoH model shall produce prediction $\hat{y}^\text{SoH}(t)$ as close as the ground-truth $y^\text{SoH}(t)$. 

\subsection{VMD Optimization for Signal Decomposition}
VMD has two key parameters that shall be optimized for achieving optimal performance of signal decomposition. The parameter optimization is often based on existing optimization algorithms, e.g., meta-heuristics. Here, we would like to find the optimal VMD parameters $K$ and $\alpha$, and meanwhile understand the performance difference of different meta-heuristics. Our VMD optimization is based on the popular PSO algorithm. We follow common settings, e.g., 100 maximum iterations, 20 particles, 0.73 for inertia weight, and 2.05 for both personal and global learning coefficients. The feasible ranges of $K$ and $\alpha$ are $[3,10]$ and $[10,2000]$, respectively. We further consider a basic genetic algorithm (GA), CSA, and grey wolf optimization (GWO) used in existing papers for parameter optimization. The fitness value convergence of the algorithms for battery B00005 is shown in Fig. \ref{fig:pso} and the results for the rest of the batteries are nearly the same.

We observe that all algorithms achieve the same optimal fitness value but with different convergence speeds. PSO converges the fastest among the tested algorithms and takes three iterations only to approximate the optimal. CSA and GWO perform similarly and take more iterations to converge and the basic GA converges the slowest with over ten iterations for minimizing the fitness value. With the same minimal fitness value, the optimal parameters produced by the tested algorithms are the same. The optimal $K$ is 3, meaning each battery capacity degradation curve shall be decomposed into three IMFs using VMD. The optimal $\alpha$ varies for different batteries, i.e., 30, 19, 92, and 10 for batteries B0005, B0006, B0007, and B0018, respectively. 

\begin{figure}
    \centering	
    \def \tmph{2.6in}
    \subfigure[VMD]{\includegraphics[height=\tmph]{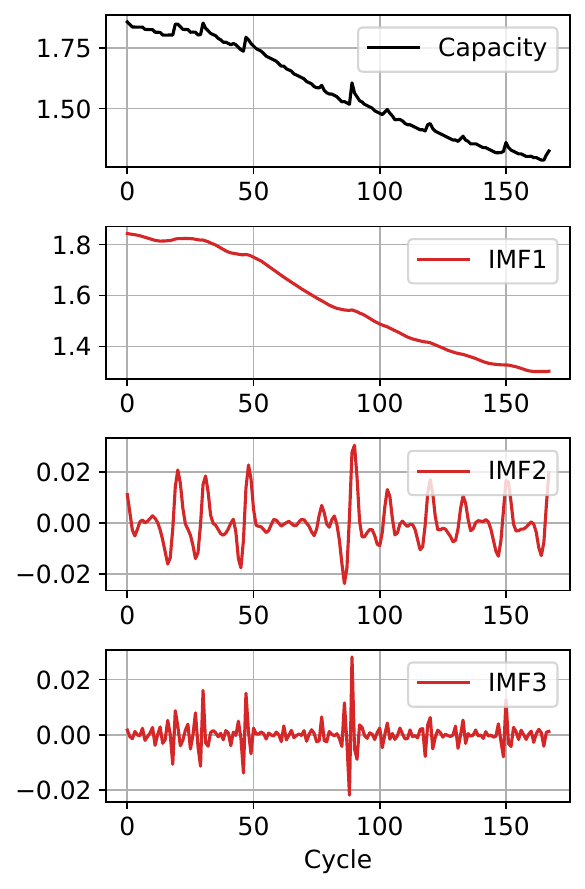}\label{fig:imf-vmd}}
    \hspace{-1em}
    \subfigure[EMD]{\includegraphics[height=\tmph]{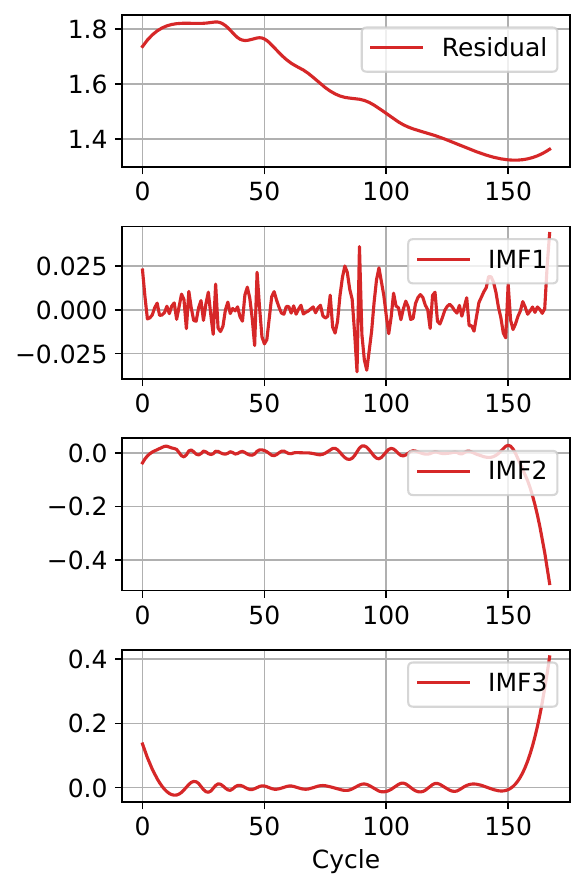}\label{fig:imf-emd}}
    \caption{Signal decomposition with VMD in Fig. \ref{fig:imf-vmd} and EMD in Fig. \ref{fig:imf-emd} for battery B0005 in Ah. VMD decomposes the original capacity curve into three IMFs for different frequency bands. EMD captures the generic aging trend with a residual curve and three IMFs without explicit frequency control.}
    \label{fig:imf}
\end{figure}

Fig. \ref{fig:imf} shows the sample $K=3$ IMFs derived by VMD and EMD for B0005. The former decomposes the original capacity degradation signal into three IMFs. IMF1 models the low-frequency battery degradation trend, and IMF3 captures high-frequency oscillations such as noises and rapid dynamic changes during charge-discharge cycles. IMF2 is mainly for intermediate frequency/variations for cycle dynamics, voltage ripples, etc. Typically, the magnitude of low-frequency values is higher than high-frequency values. For EMD, it employs a residual curve which plays a similar role as VMD's IMF1 for representing the lowest frequency trend. Such a curve can be derived by just removing the three EMD IMFs from the original capacity curve. The three IMFs for EMD are generally for high-frequency signals. Different from VMD, EMD does not introduce frequency bands, e.g., low for IMF1 and high for IMF3. Instead, the IMFs are generated sequentially where certain information might be processed repeatably without explicit frequency control, e.g., high-frequency noises reflected in several IMFs. The impact of different ways of signal decomposition will be presented in the following parts.

In summary, our results show that VMD parameters cannot be fixed for different batteries, and parameter optimization is important. However, our results discourage the exploration of different meta-heuristics for parameter optimization which offers limited performance improvement and a popular algorithm like PSO serves the optimization well enough.

\subsection{Experimental Result}
We present experimental results and discussions in this part.

\subsubsection{Experimental Setup}
We have four batteries in the dataset as set $B$. Given a specific battery $b_i \in B$ for testing and performance analysis, the remaining three batteries as $B/b_i$ are used for training and validating each ML model. In our experiments, we consider and report the performance of all four batteries. For OSL, its first convolutional layer has 128 convolutional filters with \texttt{ReLU} activation, followed by the second convolutional layer with the same settings. The layer output is flattened where all the dimensions are mapped into a one-dimensional vector for each time slot, meaning the time dimension is retained. Then, we have an LSTM layer with 64 cells and \texttt{ReLU} activation also. Finally, a dense layer maps the LSTM output to the final prediction of the battery's SoH. 

\subsubsection{Comparison Algorithms}
To showcase the performance of OSL, we introduce several comparison algorithms and report the performance achieved by them. First, we consider three algorithms without signal decomposition, including SAGPCN    \cite{10253731}, LSTM, and bidirectional LSTM (BiLSTM) \cite{Jiang08082024}. We have three signal decomposition-based algorithms. One of them, EMD-LSTM, is based on EMD and the rest two are VMD-based. The three VMD-based algorithms, including our OSL, employ LSTM, Transformer \cite{CHEN2024113388}, and CNN-LSTM, respectively. Among the seven algorithms, we get the reported results from the papers \cite{10253731}, \cite{Jiang08082024}, and \cite{CHEN2024113388} for SAGPCN, BiLSTM, and VMD-Transformer, respectively. We implemented the rest four algorithms such as LSTM and our OSL with parameter tuning. We present the experiment results in Table \ref{tab:comp}.

\begin{table}[t]
\centering
\renewcommand{\arraystretch}{1.3}
\caption{Battery health estimation performance achieved by our OSL and several comparison algorithms. The results in terms of RMSE, MAE, and MAPE, all in percentage, are reported. A small value indicates good performance. Signal decomposition is effective and OSL performs the best among the algorithms.}
\label{tab:comp}
\resizebox{0.45\textwidth}{!}{%
\begin{tabular}{c|c|c|c|c}
\hline\hline
Battery                & METHOD       & RMSE (\%) & MAE (\%) & MAPE (\%) \\ \hline
\multirow{7}{*}{B0005} 
& SAGPCN   & 0.770 & 0.550  &  $-$       \\\cline{2-5} 
& LSTM & 0.840 & 0.528 & 0.684 \\ \cline{2-5} 
& BiLSTM & 0.743     & 0.449    & $-$ \\\cline{2-5} 
& EMD-LSTM     & 2.122	& 1.838	& 2.435    \\ \cline{2-5} 
& VMD-LSTM     & 0.396	& 0.306	& 0.387   \\ \cline{2-5} 
& VMD-Trans. & $-$ & 0.248  & 0.316   \\ \cline{2-5} 
& OSL (ours) & \textbf{0.315} &	\textbf{0.207} &	\textbf{0.263}   \\ \hline\hline
\multirow{7}{*}{B0006} 
& SAGPCN & 1.650 & 1.390 &  $-$ \\ \cline{2-5} 
& LSTM & 1.213	& 0.658	& 0.828 \\\cline{2-5} 
& BiLSTM & 1.208 &	0.939 & $-$        \\\cline{2-5} 
& EMD-LSTM     & 2.342	& 2.020	& 2.747    \\\cline{2-5} 
& VMD-LSTM     & 0.522	& 0.362	& 0.469    \\ \cline{2-5} 
& VMD-Trans. & $-$ & 0.511   & 0.642  \\ \cline{2-5} 
& OSL (ours) & \textbf{0.420}	& \textbf{0.295} & \textbf{0.377}   \\ \hline\hline
\multirow{7}{*}{B0007} 
& SAGPCN  & 1.100       & 0.790     &  $-$  \\ \cline{2-5} 
& LSTM & 0.583	& 0.327	& 0.400 \\\cline{2-5} 
& BiLSTM       & 0.701     & 0.446    &  $-$  \\ \cline{2-5} 
& EMD-LSTM     & 1.112	& 0.840	& 1.028   \\ \cline{2-5} 
& VMD-LSTM     & 0.317	& 0.230	& 0.284 \\ \cline{2-5} 
& VMD-Trans. & $-$          & 0.209   & {0.252}   \\ \cline{2-5} 
& OSL (ours) & \textbf{0.288}	& \textbf{0.203}	& \textbf{0.250}    \\ \hline\hline
\multirow{7}{*}{B0018} 
& SAGPCN & 2.190  & 1.630 & $-$      \\ \cline{2-5} 
& LSTM & 1.142	& 0.706	& 0.905 \\ \cline{2-5} 
& BiLSTM   & 0.766     & 0.528    & $-$  \\ \cline{2-5}
& EMD-LSTM     & 1.452	& 1.230	& 1.637   \\ \cline{2-5} 
& VMD-LSTM     & 0.579	& 0.373	& 0.482  \\ \cline{2-5} 
& VMD-Trans. & $-$ & 0.434  & 0.559  \\ \cline{2-5} 
& OSL (ours) & \textbf{0.510}	& \textbf{0.322}	& \textbf{0.414}  \\ 
\hline\hline
\end{tabular}
}
\end{table}

\subsubsection{Results}
From the results, we have the following observations. First, OSL performs the best. It is particularly evident in battery B0005, where OSL achieves the lowest errors with 0.32\% RMSE, 0.21\% MAE, and 0.26\% MAPE. Compared to recent algorithms SAGPCN and BiLSTM without signal decomposition, the RMSE shows a significant reduction from 0.77\% and 0.74\% to OSL's 0.32\%, representing error reductions of 59\% and 58\%, respectively. OSL's performance improvement is also significant compared to the recent signal decomposition-based algorithm VMD-Transformer. Take B0006 for example, OSL's MAE and MAPE are 0.30\% and 0.38\% which are 42\% and 41\% lower than VMD-Transformer's 0.51\% and 0.64\%, respectively. Overall, OSL outperforms the comparison algorithms in all the twelve tested cases for four batteries each measured by three metrics. The consistently lower errors validate OSL's superior prediction stability, demonstrating its robust performance in handling complex battery degradation patterns.

Second, signal decomposition-based algorithms generally perform better than the rest. A direct comparison is between LSTM and VMD-LSTM. For B0005, VMD-LSTM achieves 0.40\% RMSE, 0.31\% MAE, and 0.39\% MAPE, meaning 53\%, 45\%, and 43\% error reductions compared to LSTM. BiLSTM as a more complex algorithm outperforms LSTM consistently, yet still not on par with VMD-LSTM. For example, VMD-LSTM shows 47\% and 32\% lower errors than BiLSTM for battery B0005. The effectiveness of using signal decomposition in SoH estimation might be because of the following reasons. First, with signal decomposition, noises in battery signals are decoupled from valuable battery features and ML algorithms can focus on battery health-relevant patterns to improve SoH estimation accuracy. Signal decomposition also transforms the complex capacity signals into simpler components and the simplification potentially helps ML algorithms to identify the battery health patterns effectively.

Besides, optimal battery health estimation needs effective signal decomposition. From the table, VMD is more effective than EMD. With the same LSTM, VMD-LSTM outperforms EMD-LSTM in all the test cases. For B0005, VMD-LSTM has 85\%, 89\%, and 89\% lower errors than EMD-LSTM for RMSE, MAE, and MAPE, respectively. Battery signals are often noisy and have complex patterns and overlapping frequencies. VMD helps decompose battery signals into different components corresponding to three frequency bands and serves as an effective feature extractor to improve the learning efficiency of various ML algorithms. 

\balance
ML plays a critical role also. VMD has been integrated with a few ML algorithms. The results show that a single ML model is insufficient for challenging battery analytics, and the multi-stage CNN-LSTM adopted by our OSL is more suitable. VMD-LSTM with a single LSTM model used is outperformed by OSL with CNN-LSTM in all tested cases, e.g., OSL is 20\%, 32\%, and 32\% better in terms of RMSE, MAE, and MAPE, respectively, for battery B0005. As a newer ML technology than LSTM, Transformer has demonstrated enhanced learning capability in various applications including battery analytics \cite{sameer2025ginetintegratingsequentialcontextaware}. Referring to the results reported in \cite{CHEN2024113388}, VMD-Transformer can achieve lower errors in some cases, e.g., B0007, than VMD-LSTM. However, its performance is not as competitive as OSL based on CNN-LSTM, where VMD-Transformer has a higher MAE for all batteries. We argue that a multi-stage solution with CNN-LSTM is more suitable for analyzing battery signals than single-stage solutions. Given the input battery signals, CNN first serves as a spatial feature extractor to capture local patterns such as signal peaks and fluctuations. LSTM then extracts temporal information and understands signal change patterns over time. Overall, CNN-LSTM analyzes battery signals in two stages, together capturing local and global battery signal patterns effectively, and enabling OSL to deliver accurate SoH estimations.

\section{Conclusion}
\label{sec:conclusion}
In this paper, we propose OSL, an optimal signal decomposition-based multi-stage learning for SoH estimation. OSL introduces PSO to optimize VMD parameters and reach the optimal with fast convergence. The optimized VMD is then employed to produce IMFs based on the input battery signals, where each IMF corresponds to a specific frequency band and is expected to capture different battery dynamics such as noises. We argue that the IMFs contain both spatial and temporal information that is critical for estimating SoH accurately. Thus we employ CNN in the first stage to handle spatial features and LSTM in the second stage for temporal dependencies. Extensive validation on the NASA battery dataset demonstrates the OSL's superior performance, achieving just 0.26\% MAE for the tested batteries. OSL outperforms comparison algorithms, including several latest research works, significantly. The results suggest that signal decomposition is important for batteries, but shall be optimized with effective decomposition methods. Spatiotemporal features shall also be analyzed with multi-stage learning with several ML algorithms sharing complementary strengths. Overall, OSL's ability to maintain high accuracy for SoH estimation across different battery degradation patterns makes it particularly valuable for practical battery management systems. %Future enhancements will focus on incorporating physics-informed constraints and developing physics-guided loss functions to further improve model interpretability and prediction robustness.

\bibliographystyle{IEEEtran}
\bibliography{IEEEabrv,ref_IEEE}

\end{document}